\documentclass{article}
\pdfoutput=1

\usepackage{arxiv}
\usepackage[utf8]{inputenc}             
\usepackage[T1]{fontenc}    		    
\usepackage{url}           			    
\usepackage[hidelinks]{hyperref}        
\usepackage{booktabs}       			
\usepackage{amsfonts}       			
\usepackage{nicefrac}       			
\usepackage{microtype}      			
\usepackage{lipsum}		    			
\usepackage{graphicx}
\usepackage[sort&compress, numbers]{natbib}  
\usepackage{doi}
\usepackage{subfig}
\usepackage{placeins}  					
\usepackage{multirow}
\usepackage{amsmath}
\usepackage{amssymb}
\usepackage{xcolor}

\hypersetup{
    colorlinks = true,
    linkbordercolor = white,
    linkcolor = blue,
    urlcolor  = blue,
	citecolor = blue,
}

\newcommand{\MYhref}[3][black]{\href{#2}{\color{#1}{#3}}} 	
\newcommand{\R}{\mathbb{R}}   								

\title{Reinforcement Learning Experiments and Benchmark for Solving Robotic Reaching Tasks}

\author{ \MYhref{https://orcid.org/0000-0002-9939-5537}{\includegraphics[scale=0.06]{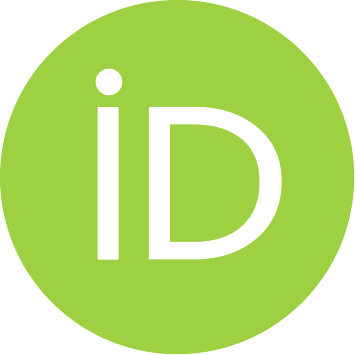}\hspace{1mm}Pierre Aumjaud}\thanks{Corresponding author} \\
	University College Dublin\\
	Dublin, Ireland \\
	\texttt{pierre.aumjaud@ucd.ie} \\
	\And
	David McAuliffe \\
	Resero Ltd \\
	Dublin, Ireland \\
	\texttt{david.mcauliffe@resero.io} \\
	\And
	\MYhref{https://orcid.org/0000-0002-8400-7079}{\includegraphics[scale=0.06]{Figures/orcid.pdf}\hspace{1mm}Francisco Javier Rodríguez Lera} \\
	Universidad de León\\
	León, Spain \\
	\texttt{fjrodl@unileon.es} \\
	\And
	\MYhref{https://orcid.org/0000-0002-4824-427X}{\includegraphics[scale=0.06]{Figures/orcid.pdf}\hspace{1mm}Philip Cardiff} \\
	University College Dublin\\
	Dublin, Ireland \\
	\texttt{philip.cardiff@ucd.ie} \\
}

\date{}

\renewcommand{\headeright}{Preprint}
\renewcommand{\undertitle}{}
\renewcommand{\shorttitle}{Reinforcement Learning Benchmark for Robotics Tasks}

\hypersetup{
pdftitle={Reinforcement Learning Experiments and Benchmark for Solving Robotic Reaching Tasks},
pdfsubject={cs.LG, stat.ML},
pdfauthor={Pierre Aumjaud, David McAuliffe, Francisco Javier Rodríguez Lera, Philip Cardiff}, 
pdfkeywords={Reinforcement learning, Robotics, Benchmark, Model-free, Sim-to-real},
}

\begin{document}
\maketitle

\begin{abstract}
Reinforcement learning has shown great promise in robotics thanks to its ability to develop efficient robotic control procedures through self-training. In particular, reinforcement learning has been successfully applied to solving the reaching task with robotic arms. In this paper, we define a robust, reproducible and systematic experimental procedure to compare the performance of various model-free algorithms at solving this task. The policies are trained in simulation and are then transferred to a physical robotic manipulator. It is shown that augmenting the reward signal with the Hindsight Experience Replay exploration technique increases the average return of off-policy agents between 7 and 9 folds when the target position is initialised randomly at the beginning of each episode.
\end{abstract}

\keywords{Reinforcement learning \and Robotics \and Benchmark \and Model-free \and Sim-to-real}

\section{Introduction}

Reinforcement learning (RL) is a paradigm in the field of machine learning that has recently gained tremendous interest \cite{Sutton2018}. Unlike supervised learning, RL is capable of learning sequential decision-making policies by interacting with an environment, while sometimes achieving super-human performance \cite{Silver2016}. RL requires a reward function to be defined, which is why it has naturally been applied to games or toy problems \cite{Mnih2015}. However, real-world applications are still scarce and challenging. Robotics allow researchers to define controlled training environments relatively easily, thus RL has found many successful applications in this field. 

Reinforcement learning approaches can be grouped in two categories: model-based -- where the agent first learns a model of the environment and then uses it to make predictions; and model-free -- where the agent directly learns a policy based only on its interactions with the environment. Historically, model-based RL has first been applied to robotics \cite{Deisenroth_k} due to its high sample efficiency, allowing agents to solve tasks with a limited number of policy iterations. Nevertheless, it is sometimes challenging to build accurate models for complex robotics problems, causing model-based approaches to often suffer from poor asymptotic performance. The invention of Deep Deterministic Policy Gradients \cite{Lillicrap2015} has paved the way for other model-free RL algorithms to successfully solve problems with continuous state and action space. Since then, considerable improvements have occurred in the field with the invention of Twin Delayed Deep Deterministic Policy Gradient \cite{Fujimoto2018}, Soft Actor Critic \cite{Haarnoja2018, Haarnoja2018a} and Hindsight Experience Replay \cite{Andrychowicz2017}, which addresses the problem of sparse rewards.

One of the most common robotic tasks studied in an RL context is trajectory planning to reach a target position. A number of approaches have been adopted to solve this problem, including model-based approaches \cite{Devin2017a, Gupta2017}, model-free approaches \cite{Plappert2018, Chen2018, RupamMahmood2018, Tavakoli2019, Gu2016, Luo2020, Pham2018, Lucchi2020}, a combination of both \cite{Pong2018}, or learning acceleration via closed-loop policies \cite{Pinto2018}. 

In this article, we will consider solving the problem of reaching target positions both using a robotic simulator and a physical manipulator. The equipment, training environments and methods adopted to perform the performance benchmark of a number of model-free RL algorithms are described in Section \ref{methods}. The comparative results are reported and discussed in Section \ref{results}, both in terms of training convergence and evaluation metrics. Finally, a general conclusion is given and directions for future works are outlined in Section \ref{conclusion}.

\section{Environments and Methods} \label{methods}

\subsection{Manipulator Description}

\begin{figure*}[!htbp]
\centering
\subfloat[Virtual environment]{
\includegraphics[width=0.48\textwidth]{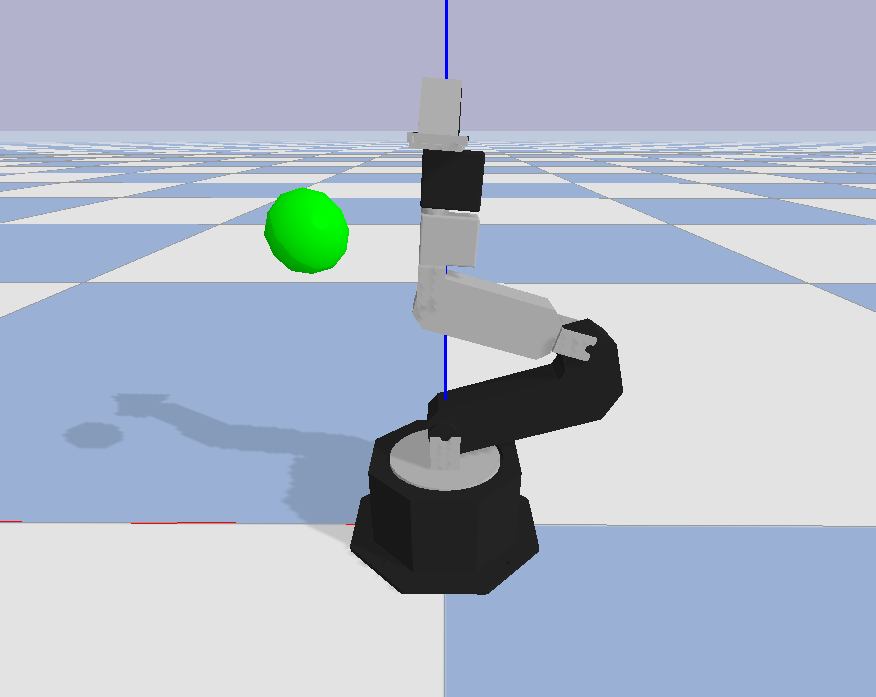}
\label{fig:sim_env}}
\hfil
\subfloat[Physical environment]{
\includegraphics[width=0.48\textwidth]{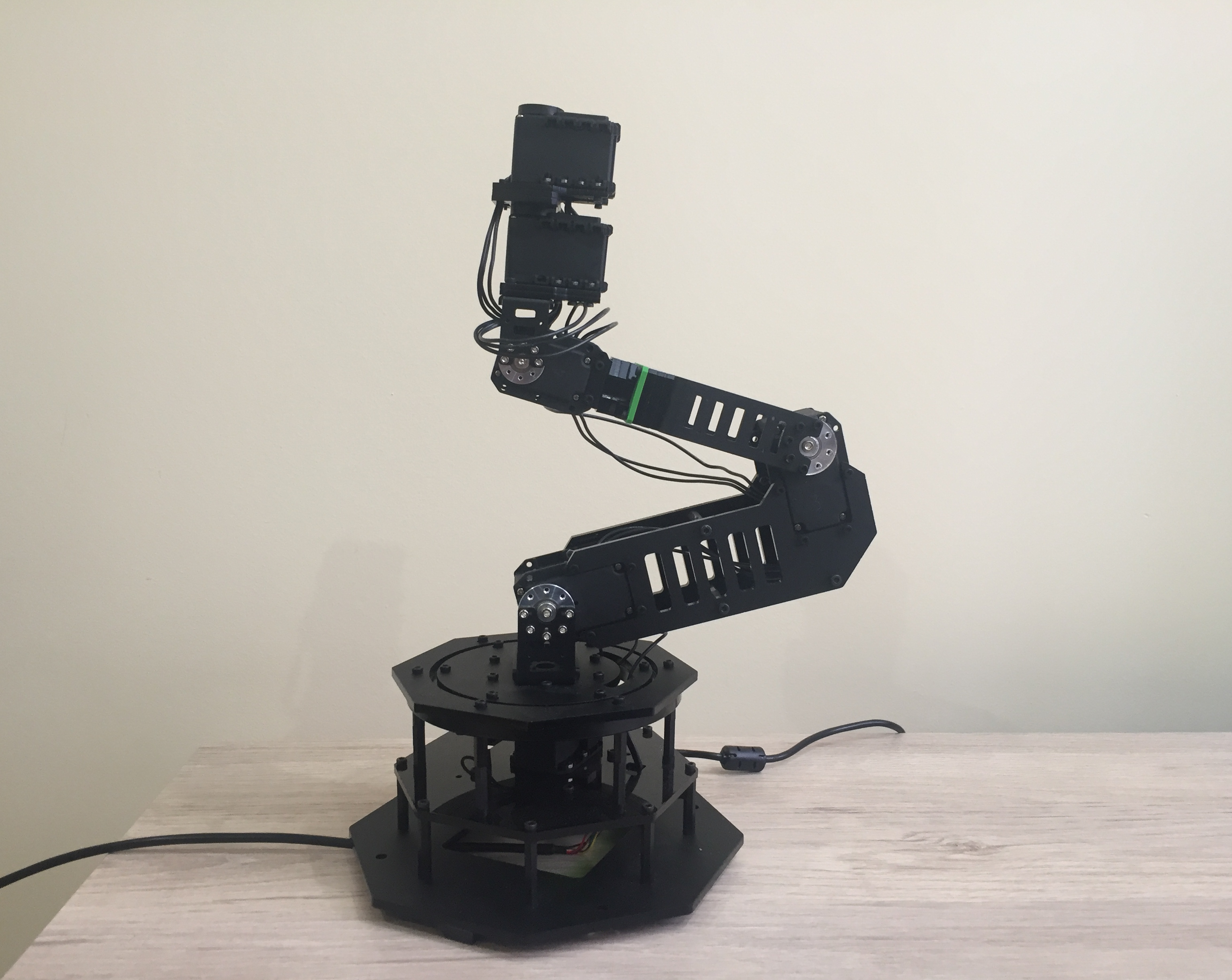}
\label{fig:phys_env}} 
\hfil
\caption{The simulated (Pybullet) and physical (WidowX MKII manipulator) learning environments in their initial episode position. The target position is indicated by a green sphere.}
\label{fig:environment}
\end{figure*}

The robotic arm used in this work is the WidowX MKII manipulator by Trossen Robotics \cite{TrossenRobotics}, see Fig. \ref{fig:phys_env}. It is a 6-joints robotic arm possessing an 82 cm diameter reach and equipped with Dynamixel servo actuators and parallel grippers. The robot arm is compatible with the Robot Operating System (ROS) \cite{Quigley09} and is shipped with its associated ROS packages. The Replab project \cite{Yang2019} has previously employed the same robot arm to solve grasping tasks and provide a reproducible benchmark platform for robotics learning. The learning environments used here are adapted from those in the Replab project. The Pybullet physics engine \cite{Coumans} simulates the response of the robot arm in a virtual environment, which accelerates the training process compared to training on the physical robot alone.

\subsection{Reaching Task and Environment Definition}

The reaching task considered here is a well-known problem in the robotic manipulation field. It consists of determining the optimal sequence of actions required to bring the robot arm's end effector as close as possible to given target coordinates within its workspace. 

In order to solve the reaching task in an RL context, the problem is formalised as a Markov Decision Process. The environment's \textit{state} $s_t$ is defined as an array holding the current spatial coordinates of the robot's end effector and the angular position of each joint:

\begin{equation}
\forall	s_t \in S, \qquad  s_t = [x_e, y_e, z_e, \theta_1, \theta_2, \theta_3, \theta_4, \theta_5, \theta_6]_t  
\end{equation}

where $(x_e, y_e, z_e)$ are the Cartesian coordinates of the end effector and $\theta_i$ is the angular position of joint $i$ in radians. The \textit{actions} $a_t$ sent to the environment are defined as an array holding the changes in joint angle from the previous position:

\begin{equation}
\forall a_t \in A,  \qquad a_t = [\delta_1, \delta_2, \delta_3, \delta_4, \delta_5, \delta_6]_t 
\end{equation}

where $\delta_i$ is the change in angular position from the previous position of joint $i$. $S$ and $A$ denote the set of possible states and actions, respectively. These spaces are finite and continuous in this context. The \textit{reward} $r_t$ is defined as the negative, squared, L2 norm of the vector between the current end effector position and the goal position (dense reward setting):

\begin{equation}
\forall r_t \in \R^-, \qquad r_t = - [(x_e - x_g)^2 + (y_e - y_g)^2 + (z_e - z_g)^2]_t
\end{equation}

where $(x_g, y_g, z_g)$ are the Cartesian coordinates of the goal pose. Each learning iteration is composed of an action sent, a state and a reward received by the agent, which defines a \textit{timestep} $t$. A sequence of 100 timesteps defines one \textit{episode}, unless the termination condition -- the episode ends if the distance between the end effector and the goal is less than 0.5 mm -- is reached before this. The sum of the rewards over one episode defines the \textit{return} $R_t$, see equation below.

\begin{equation}
R_t = \sum_{t=1}^{100} r_t
\end{equation}

The \textit{policy} maps the environment's current state to the next action with the objective of maximising the return. A deterministic policy is a function defined as $\pi_d : S \mapsto A $ and a stochastic policy is a probability distribution defined as $\pi_s(A | S)$, meaning that the probability of selecting an action depends on the current state.

The robot arm joints are always initialised with the same angle positions at the beginning of each episode both in the virtual and the physical environments, as shown in Fig. \ref{fig:environment}. The goal position is either constant across all training and evaluation episodes (fixed goal) or initialised randomly at the beginning of each episode (random goal). Either way, the goal position is always within the reach of the robot arm.

The learning environments are implemented with OpenAI Gym \cite{Brockman2016}, a standard toolkit for developing and comparing reinforcement learning algorithms. A goal-oriented Gym environment \cite{Brockman2016} is also implemented in order to explore the observation space more efficiently using Hindsight Experience Replay. In this case, the desired goal and achieved goal coordinates are also included in the state definition. In total, four learning environments are implemented:

\begin{itemize}
  \item \textbf{Env1}: Pybullet simulation + fixed goal
  \item \textbf{Env2}: Pybullet simulation + random goal
  \item \textbf{Env3}: Physical robot + fixed goal
  \item \textbf{Env4}: Physical robot + random goal
\end{itemize}

The policies are trained on the Pybullet environments only (i.e. Env1 and Env2), however they are evaluated both in simulation and on the physical robot.

\subsection{Experiments and Benchmark} 

It is proposed to compare the performance of a number of RL algorithms at solving the reaching task using the environments described above. The RL algorithms considered in the benchmark are: 
\begin{itemize}
\item Advantage Actor Critic (A2C) \cite{Mnih2016}
\item Actor Critic using Kronecker-Factored Trust Region (ACKTR) \cite{Wu2017}
\item Deep Deterministic Policy Gradients (DDPG) \cite{Lillicrap2015}
\item Proximal Policy Optimization (PPO) \cite{Schulman2017}
\item Twin Delayed Deep Deterministic Policy Gradient (TD3) \cite{Fujimoto2018}
\item Soft Actor Critic (SAC) \cite{Haarnoja2018, Haarnoja2018a}
\item Trust Region Policy Optimization (TRPO) \cite{Schulman2015a}
\item SAC + Hindsight Experience Replay (HER) \cite{Andrychowicz2017}
\item TD3 + Hindsight Experience Replay \cite{Andrychowicz2017}
\end{itemize}
  
All these algorithms are compatible with continuous state and action spaces environments, which is a requirement for the reaching task considered here. DDPG, SAC, TD3 and HER learn a deterministic policy and a stochastic policy is learnt by the rest of the algorithms. The two off-policy algorithms -- SAC and TD3 -- are combined with HER in an effort to reduce sample complexity. The algorithms are applied using the Stable Baselines implementation \cite{Hill2018}.

The hyperparameters of each algorithm are first optimised with the Optuna framework \cite{Akiba2019} over 100 training runs of 100 episodes each. In each run, the hyperparameters are sampled with the Tree-structured Parzen Estimator and unpromising runs are stopped early using a median pruner. Subsequently, RL agents are trained in the virtual environments Env1 and Env2 for 200,000 timesteps using each algorithm described above. The A2C, ACKTR and PPO agents are trained over 8 parallel environments in order to speed up the learning process (the other algorithm's implementations do not currently support parallelisation). The training is performed on University College Dublin's Sonic HPC cluster \cite{UniversityCollegeDublin}, which is equipped of two NVIDIA Tesla V100 16 GB GPUs per node. The training is repeated over 10 different initialisation seeds and the metrics are averaged in order to strengthen the robustness of the experiments. Once the optimal policies are learnt in the virtual environments, they are deployed and tested both in the virtual (Env1 and Env2) and physical environments (Env3 and Env4). All policies are evaluated with their associated agent for 100 episodes. A random policy is also evaluated to serve as a reference. In the case of physical environments, the model weights of the best-performing seed run (in terms of return) are transferred to the agent controlling the physical robot. In the case of the virtual environment, all seed runs are evaluated and the metrics are averaged. The following evaluation metrics are reported:

\begin{itemize}
  \item \textbf{Average return}: Return averaged over the 100 evaluation episodes.
  \item \textbf{Train time}: Average training time over the 10 random seeds in minutes.
  \item \textbf{Success ratio @X mm}: number of episodes where the end effector ended within a distance threshold of X mm from the goal position, divided by the total number of evaluation episodes. The distance thresholds considered are 50 mm, 20 mm, 10 mm and 5 mm.
  \item \textbf{Reach time @X mm}: number of timesteps required to reach the target during a successful episode within a distance threshold of X mm. 
\end{itemize}

The source code repository and an explanatory video are available at \url{https://git.io/JJJdu} and \\ \url{https://youtu.be/-N-6Me8UkFk}.

\section{Results and Discussion}  \label{results}

\subsection{Training Convergence}

\begin{figure*}[!htbp]
\centering
\includegraphics[width=0.7\linewidth]{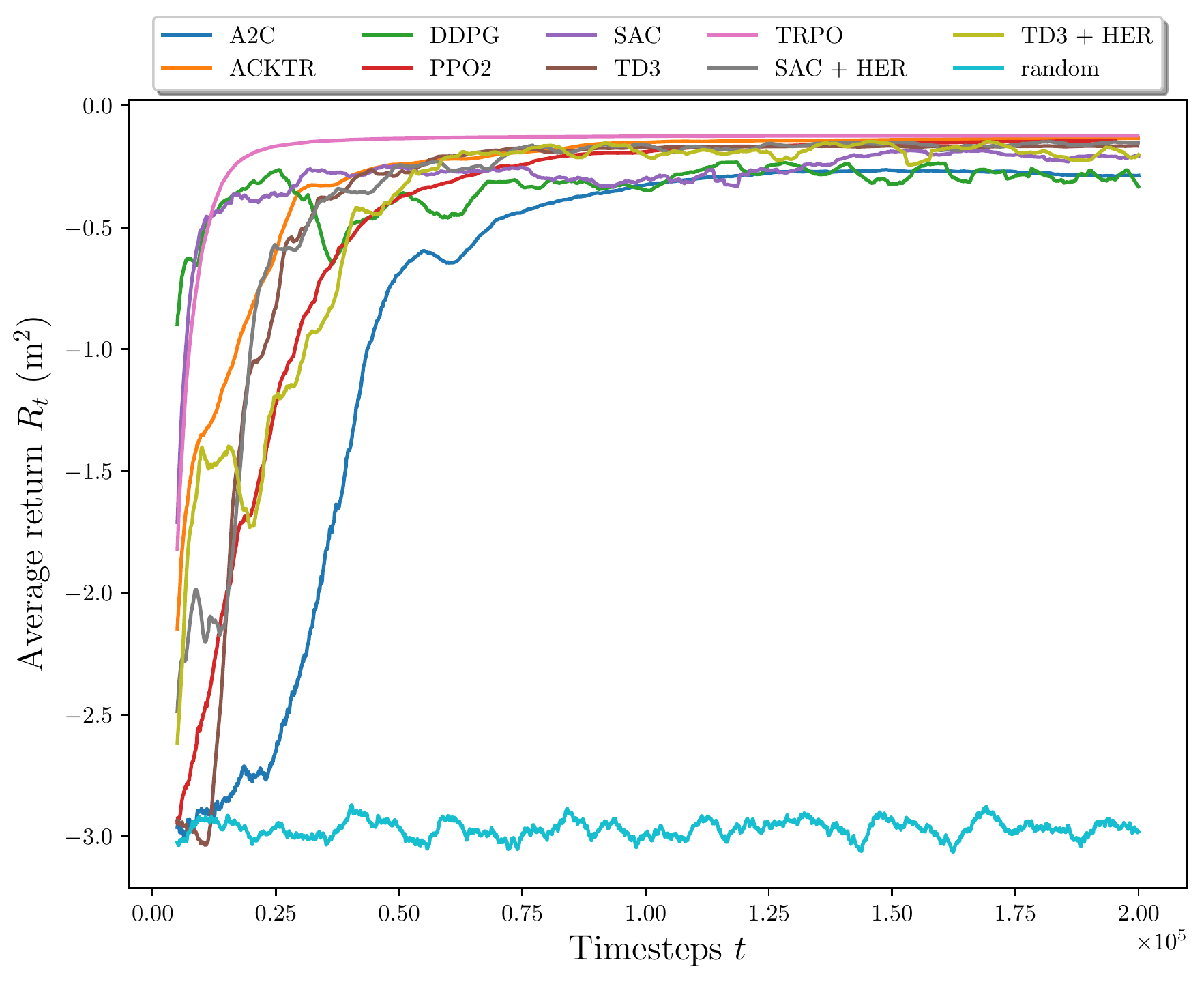}
\caption{Convergence curves of the algorithms solving Env1 (fixed goal)}
\label{fig:learning_curves_env1}
\end{figure*}

\begin{figure*}[]
\centering
\includegraphics[width=0.7\linewidth]{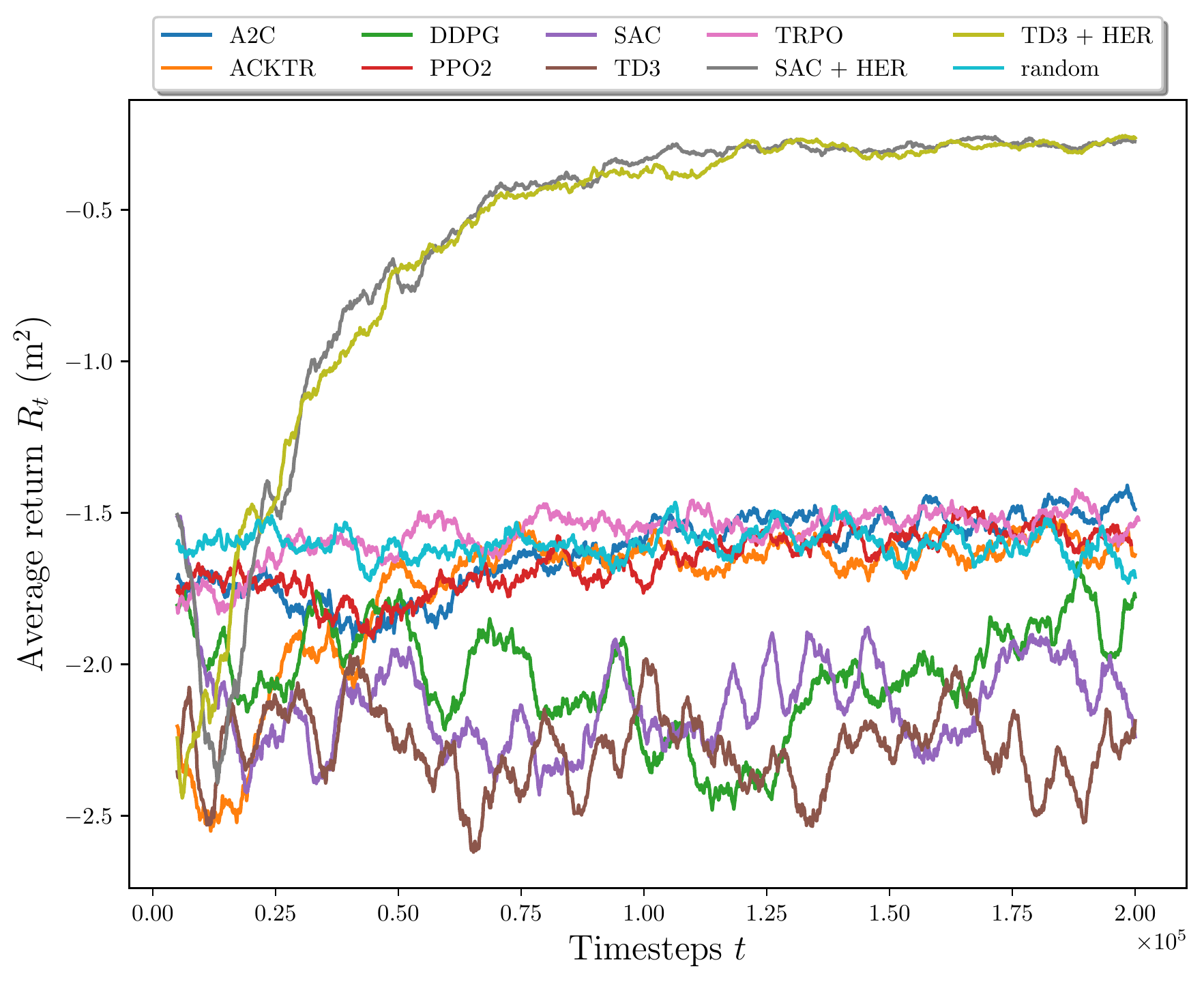}
\caption{Convergence curves of the algorithms solving Env2 (random goal)}
\label{fig:learning_curves_env2}
\end{figure*}

\begin{figure*}[!htbp]
\centering
\subfloat[A2C]{\includegraphics[width=0.32\textwidth]{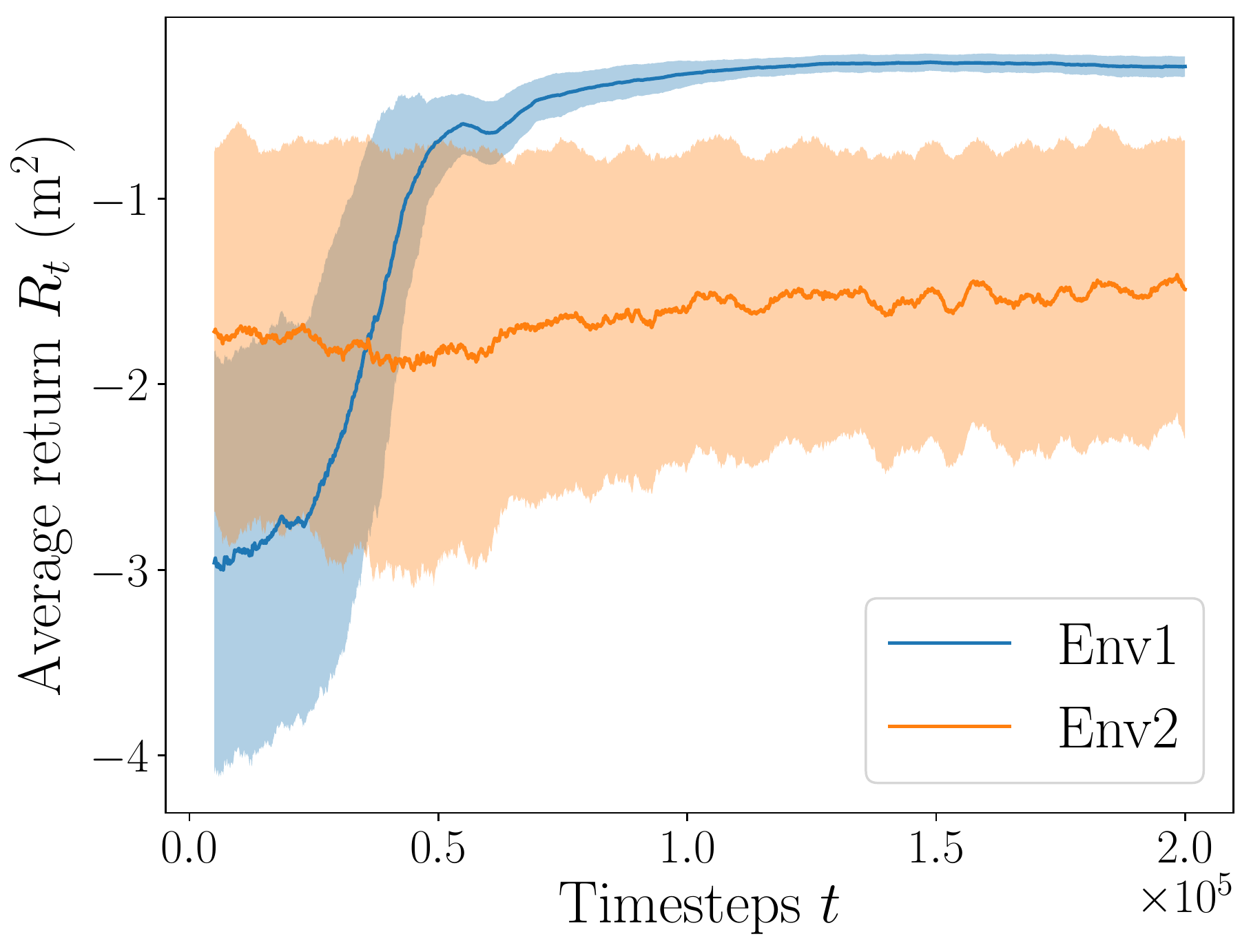}
\label{fig:a2c}}
\hfil
\subfloat[ACKTR]{\includegraphics[width=0.32\textwidth]{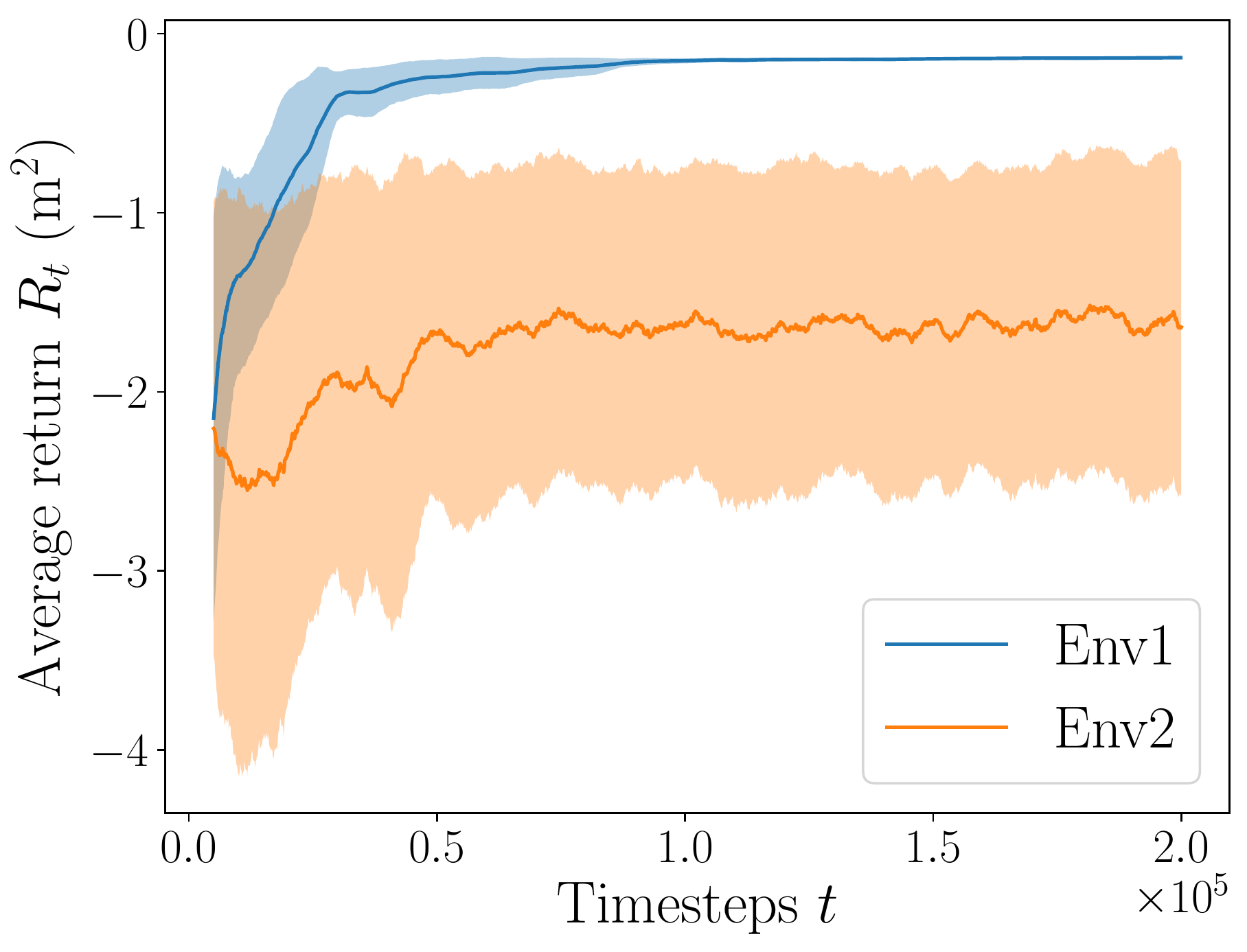}
\label{fig:acktr}} 
\hfil
\subfloat[DDPG]{\includegraphics[width=0.32\textwidth]{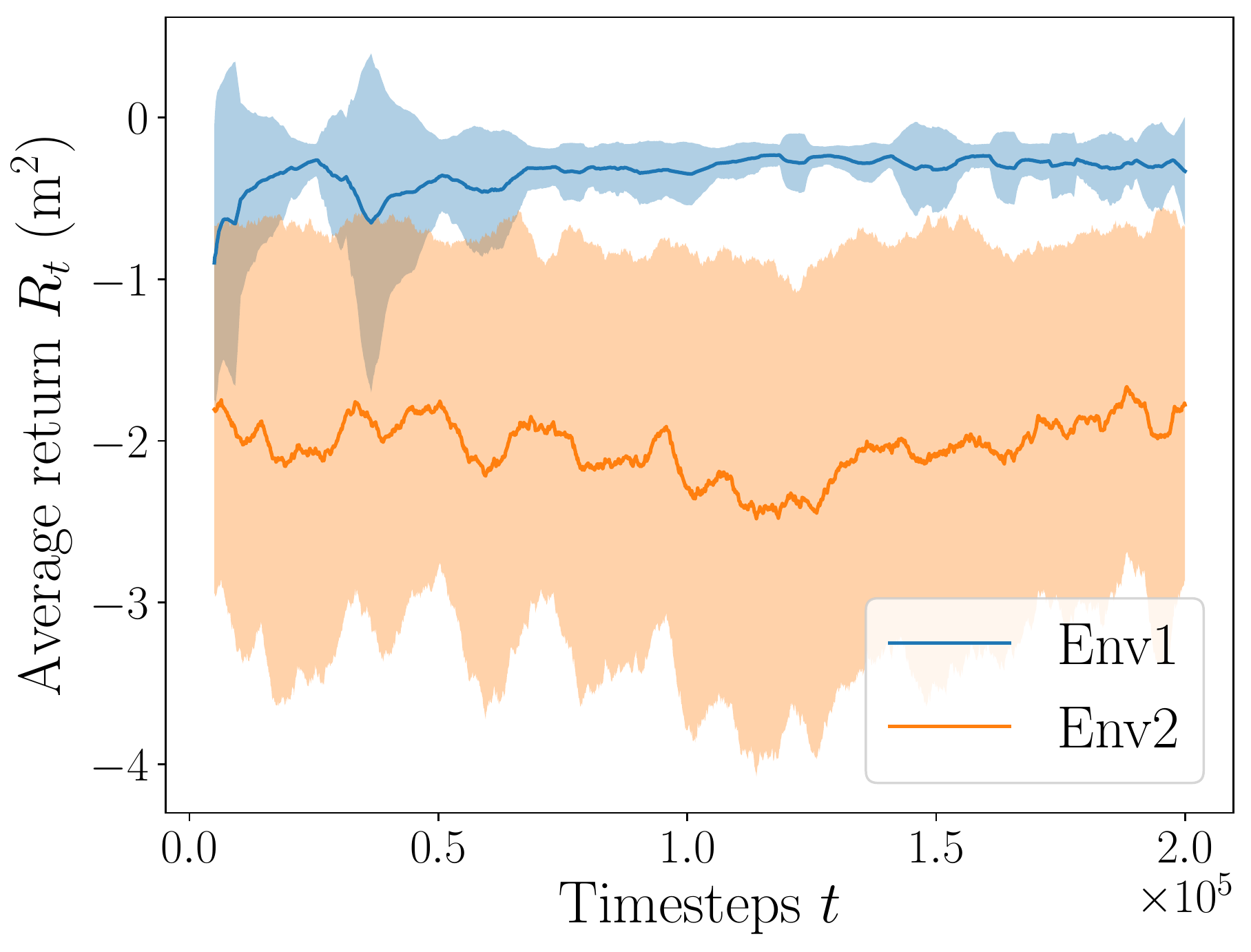}
\label{fig:ddpg}} 
\hfil
\subfloat[PPO]{\includegraphics[width=0.32\textwidth]{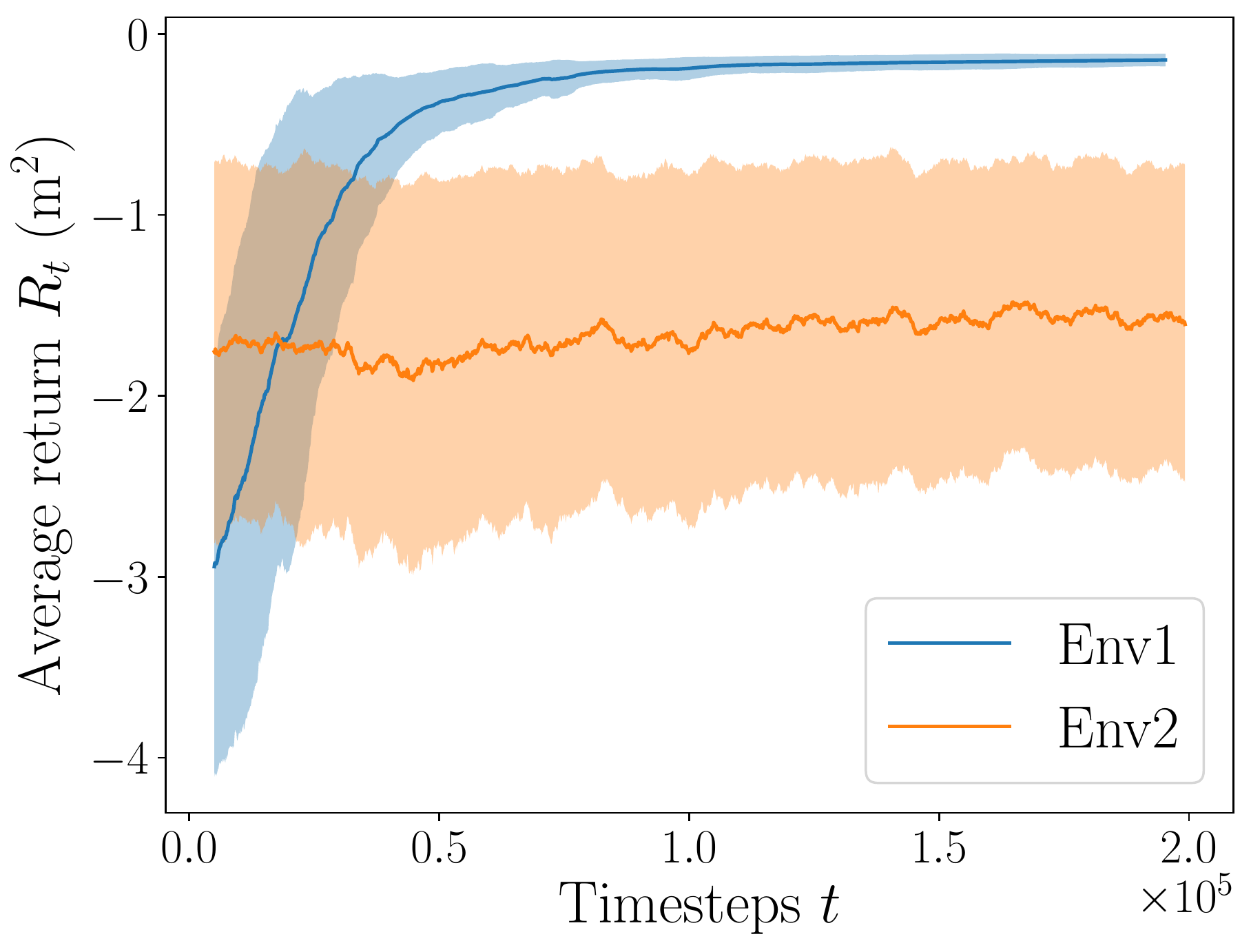}
\label{fig:ppo}} 
\hfil
\subfloat[SAC]{\includegraphics[width=0.32\textwidth]{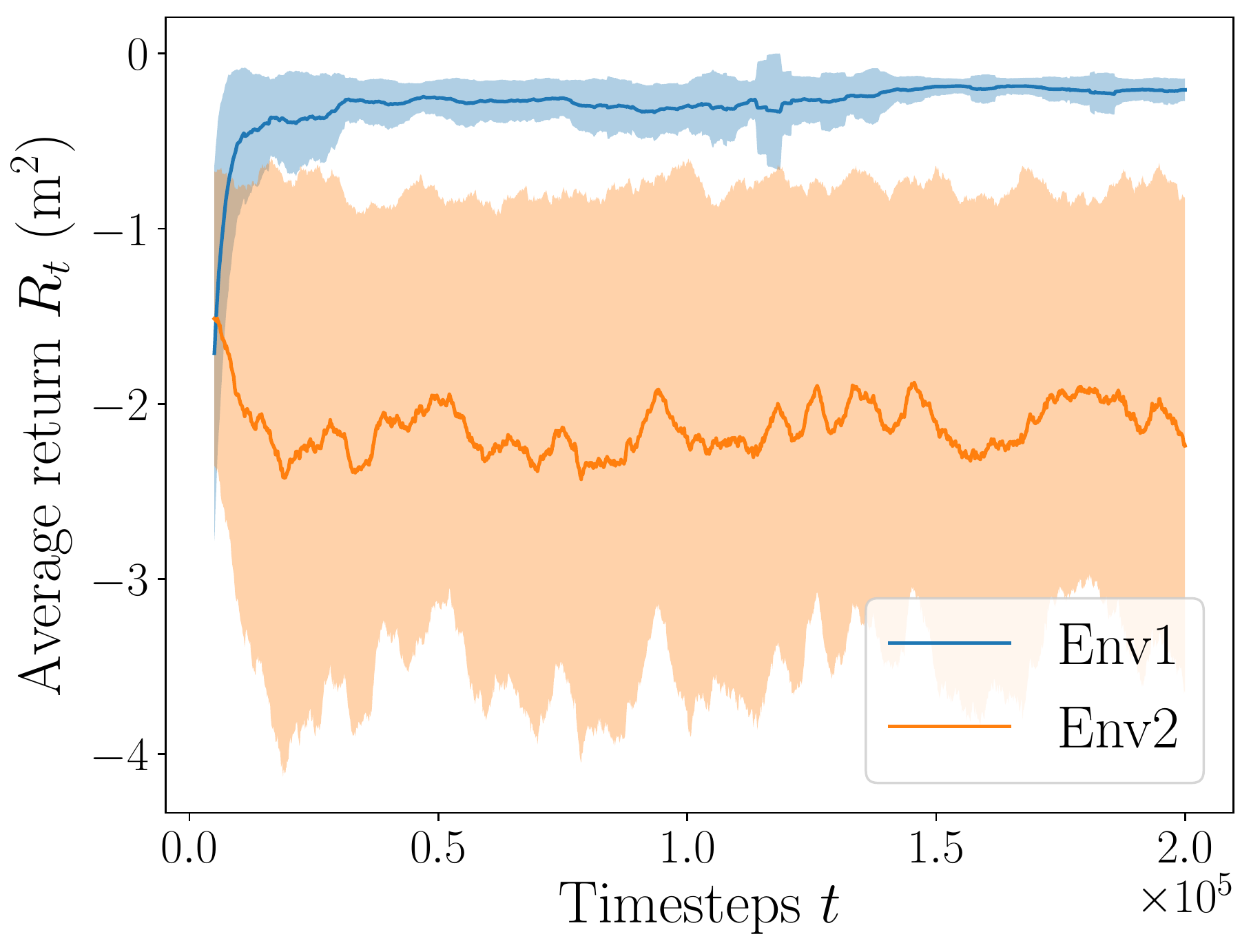}
\label{fig:sac}} 
\hfil
\subfloat[TD3]{\includegraphics[width=0.32\textwidth]{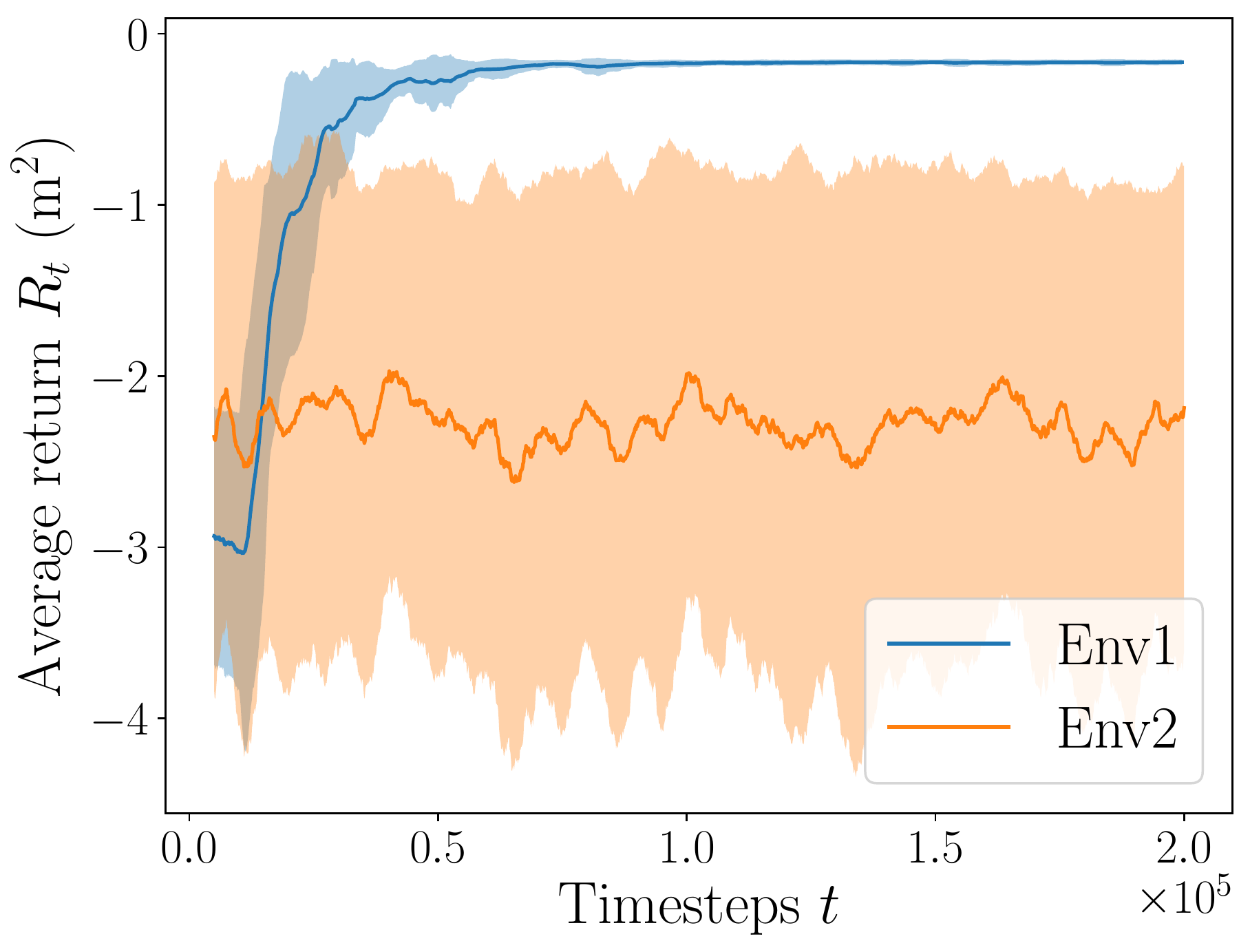}
\label{fig:td3}} 
\hfil
\subfloat[TRPO]{\includegraphics[width=0.32\textwidth]{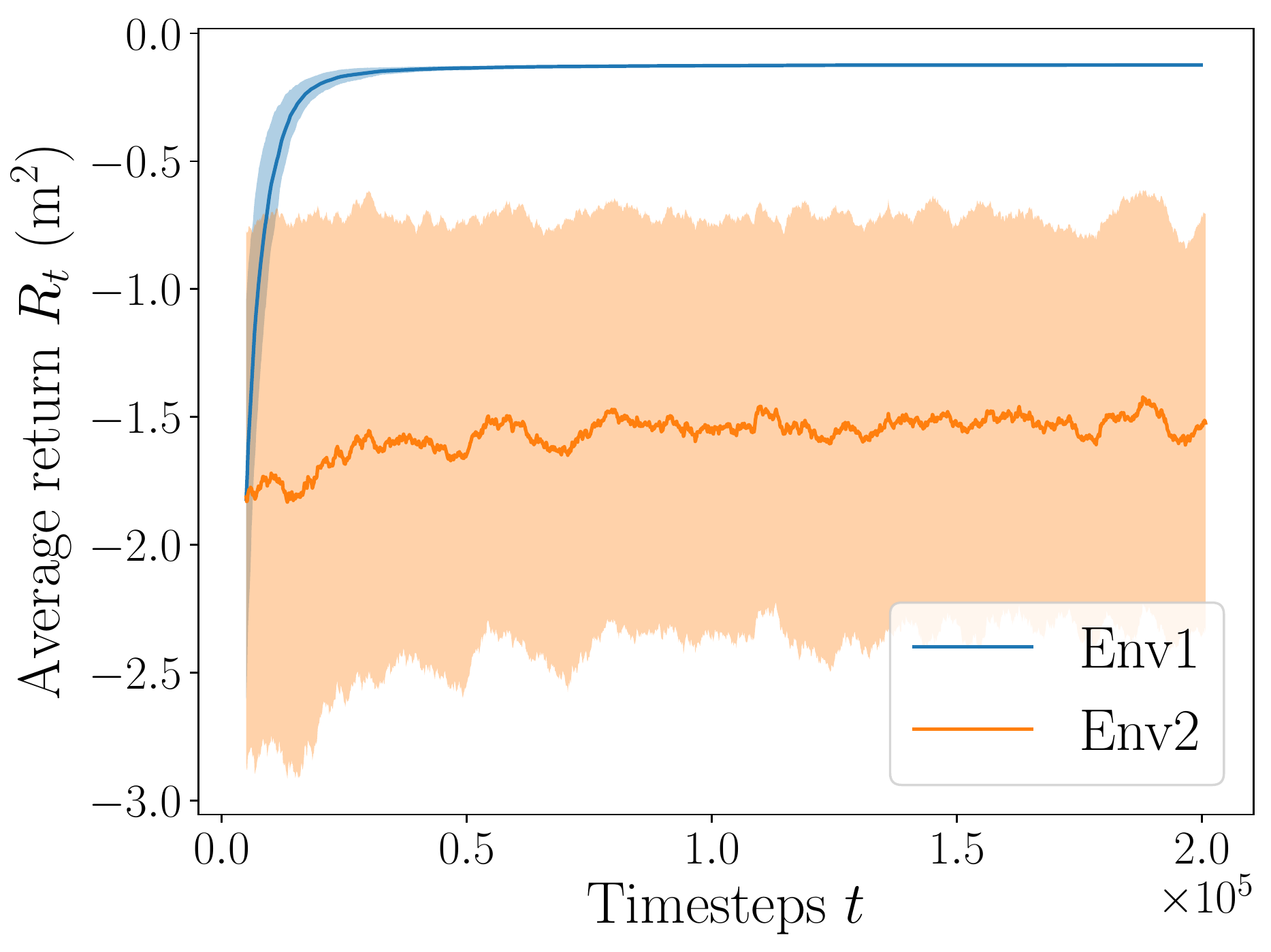}
\label{fig:trpo}} 
\hfil
\subfloat[HER + SAC]{\includegraphics[width=0.32\textwidth]{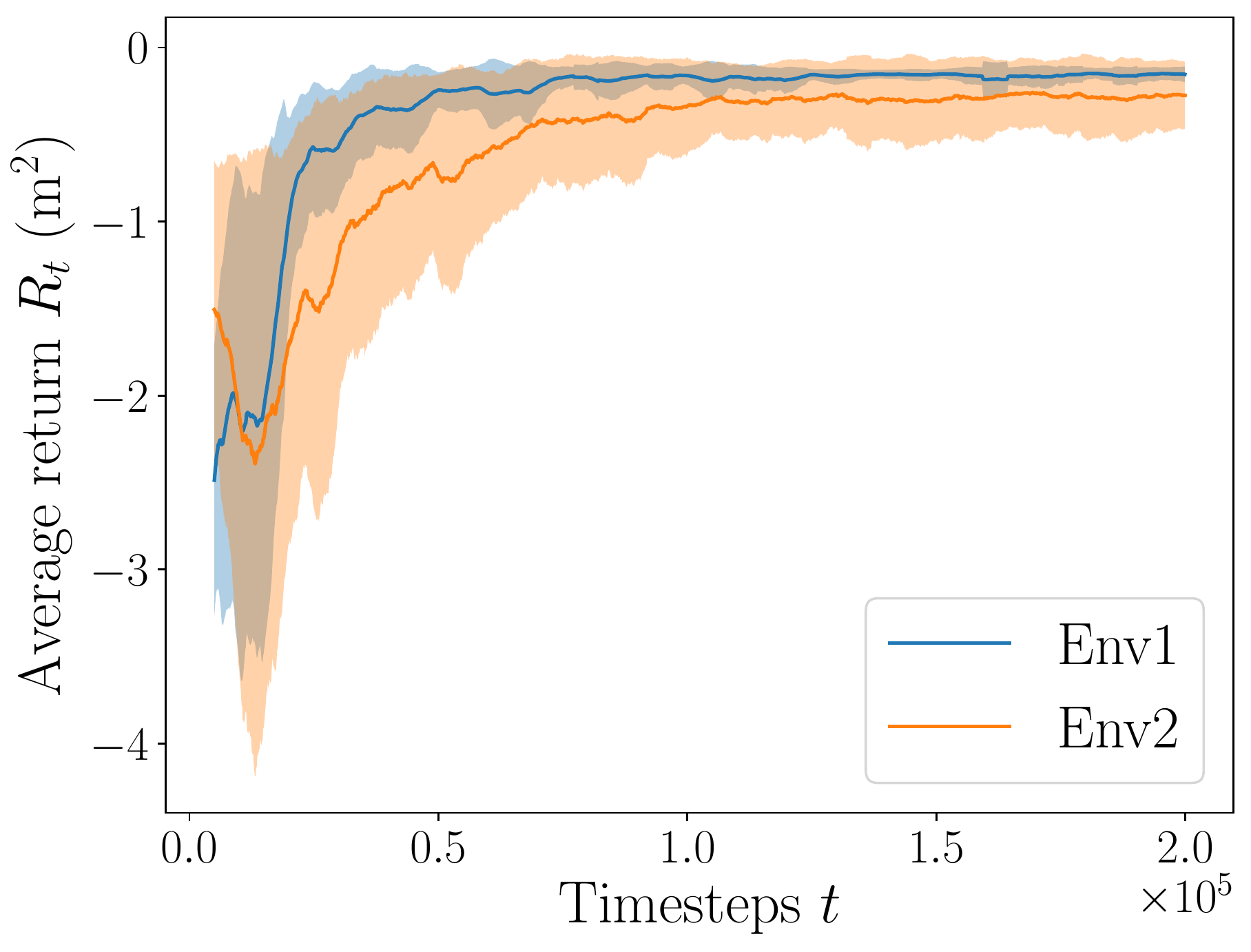}
\label{fig:her_sac}} 
\hfil
\subfloat[HER + TD3]{\includegraphics[width=0.32\textwidth]{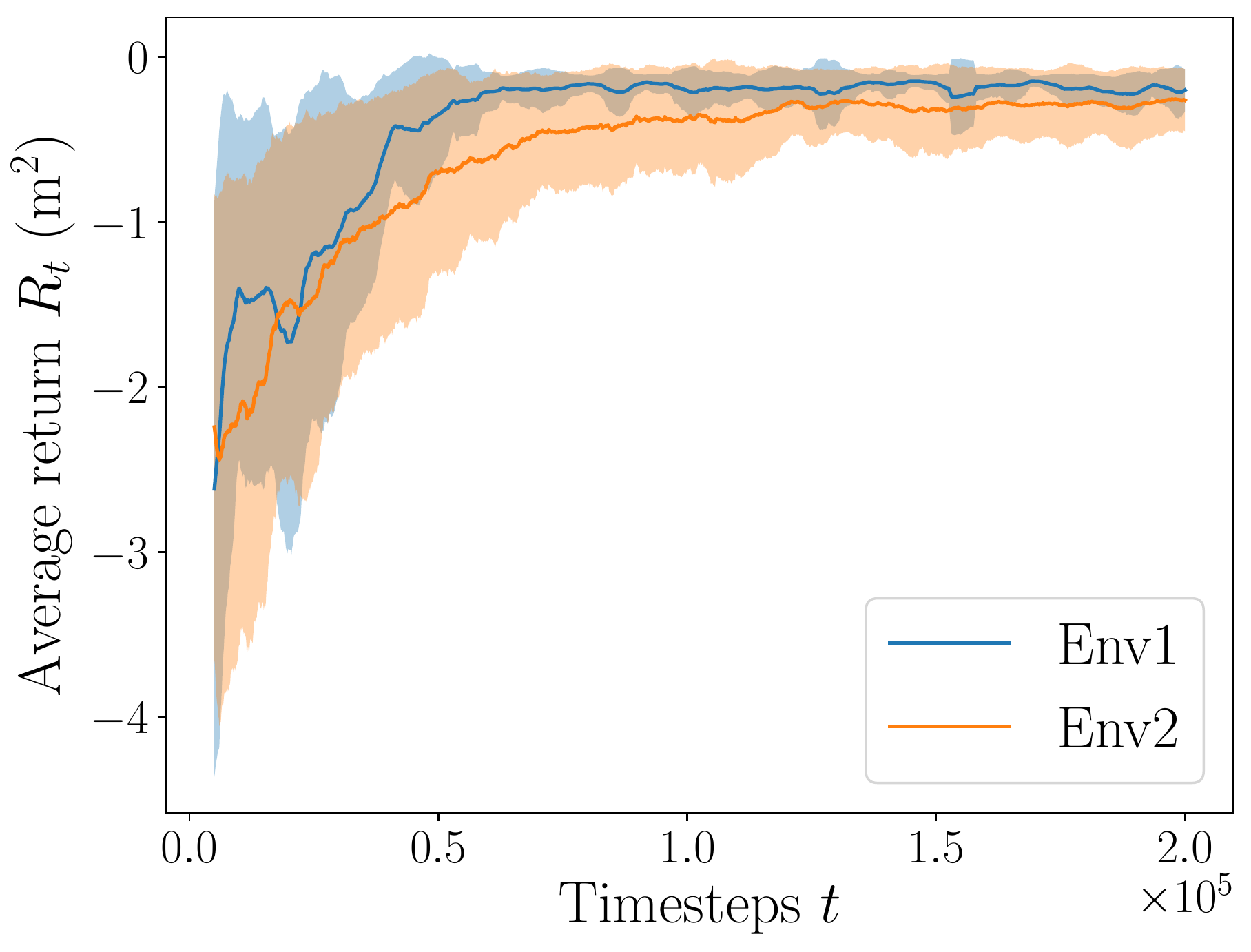}
\label{fig:her_td3}}
\hfil
\caption{Convergence curve comparison between Env1 (fixed goal -- in blue) and Env2 (goal initialised randomly -- in orange)}
\label{fig:training_curves_comp}
\end{figure*}

The training convergence curves averaged over the 10 initialisation seeds are shown in Figs. \ref{fig:learning_curves_env1} and \ref{fig:learning_curves_env2} for the two virtual environments considered, Env1 (fixed goal) and Env2 (random goal). The curves are smoothed with a rolling window average of 50 for clarity purposes. A comparison of the convergence curve between Env1 and Env2 is also reported for each algorithm individually, see Fig. \ref{fig:training_curves_comp}. In these plots, the middle line represents the average reward over the 10 seed runs at each timestep and a shaded area is drawn at $\pm$ the standard deviation.

It can be noted that Env2 (random goal) is a more challenging task to solve since the average return is generally lower than that of Env1 (fixed goal) at the end of the training process. All algorithms manage to learn a successful policy in Env1 after 200,000 timesteps. DDPG, TRPO and SAC achieve the highest sample efficiency as their learning curves plateau the earliest after less than 20,000 timesteps, whereas A2C exhibits the highest sample complexity. Moreover, TRPO, TD3 and ACKTR produce the most stable and repeatable training as their standard deviations are the lowest at the end of training, see Fig. \ref{fig:training_curves_comp}.

When the goal is initialised randomly at the beginning of the episode, most algorithms fail to learn a successful policy. Some algorithms perform even worse than a random policy, for example DDPG in Env2. However, combining SAC and TD3 with HER yields a healthy training, even in Env2, as illustrated in Fig. \ref{fig:learning_curves_env2}. Accelerating the observation space exploration this way tends to reduce slightly the sample efficiency and the training stability of the off-policy algorithms SAC and TD3, see Figs.  \ref{fig:her_sac}  and \ref{fig:her_td3}.

\subsection{Evaluation Metrics}

Although the training convergence curve can be a good proxy to measure the performance of an RL agent, it is more precise to consider evaluation metrics. The evaluation metrics of all trained agents and all four environments considered are reported in Tables \ref{tab:raw_results} and \ref{tab:raw_results2}. The metrics are averaged over 10 trainings initialised with different seeds; the standard deviations are not reported here for clarity but are available upon request or in the associated source code repository.

When the goal position is fixed, the average return of successful agents is much higher than that of the random policy, which agrees with the observations from the convergence curves, see for example TRPO in Env1. In this case, the best-performing algorithms in terms of return are TRPO and TD3. When the goal position is initialised randomly however, only the algorithms combined with HER achieve an average return significantly higher than random.

The training walltime is substantially shorter for algorithms supporting parallelisation, as expected. For instance with a train walltime of 68 seconds, PPO is almost 19 times faster to train than SAC + HER in Env2. It is also about twice as fast as a random policy that does not exploit any deep learning architectures nor uses parallel environments.

The success ratio measures the accuracy of an agent at reaching the target and the reach time measures their speed at reaching the target during an episode (note that the maximum reach time is equal to the length of an episode, 100).

In the least challenging environment -- Env1 (fixed goal), all algorithms achieve a high success ratio and a fast reach time when the distance threshold is large (i.e. 50 mm). As expected, the agent's accuracy decreases as the distance threshold is becoming tighter, and it drops dramatically beyond a distance threshold of 20 mm. In this situation, TD3 is the most accurate algorithm with a success ratio @5 mm of 0.5. The agents capable of reaching their target quickly are generally those exhibiting a high success ratio and average return. The fastest agents are TD3 and TRPO with a reach time of around 9 timesteps at the 50 mm threshold, although the other successful agents are only slightly slower.

When the goal is initialised randomly (Env2), fewer agents managed to learn a successful policy than in Env1, which is reflected by the much lower average return. The best-performing agents are those leveraging the HER exploration strategy, i.e. SAC + HER and TD3 + HER with a success ratio @50 mm of 0.67 and 0.64, respectively. This phenomenon was also observed in the training convergence, see Fig. \ref{fig:training_curves_comp}. Similar to Env1, the success ratio diminishes rapidly as the distance threshold reduces. The negligible success ratios @50 mm observed in the other algorithms are comparable to that of a random policy, meaning that the agent has occasionally and fortuitously reached the target.

A sharp decrease in performance is globally observed when trained policies are transferred to the physical robot and evaluated in Env3 and Env4, both in terms of average return and success ratio. This could be further improved by continuing the training from the virtual environments in the physical environments, as the Pybullet simulation does not take into account the joint friction and other imperfections of the physical manipulator. Such transfer from a simulated to the physical world is a well-known challenge in robotics, often referred to as sim-to-real.

However, it is worth noting that PPO, TD3 and TD3 + HER achieved a relatively successful policy transfer from the simulated (Env1) to the physical environment (Env3) as exhibited by their success ratio @50 mm of 1 and a comparable reach time. Finally in the case of the physical environment with random goal (Env4), the only successful policies are those combined with HER. This was also noted in the associated virtual environment with random goal, Env2..

\begin{table*}[!htbp]
\centering
\caption{Evaluation metrics of all trained agents for the four environments}
\begin{tabular}{p{22mm}|p{22mm}|p{14mm}p{17mm}p{14mm}p{14mm}p{14mm}p{14mm}}
\toprule
Environment & Algorithm & Average return $R_t$ (m\textsuperscript{2}) & Average train walltime (s) & Success ratio @50 mm & Reach time @50 mm & Success ratio @20 mm & Reach time @20 mm \\
\midrule
\multirow{10}{*}{\shortstack[l]{Env1  \\ (simulation \\ + fixed goal)}}      
& A2C           & -0.30  & 55.0   & 0.94   & 18.01 & 0.26 & 37.0 \\
& ACKTR         & -0.13  & 55.3   & 1.00   & 9.26  & 0.99 & 13.5 \\
& DDPG          & -0.29  & 635.0  & 0.90   & 9.2   & 0.30 & 12.3 \\
& PPO           & -0.14  & 65.0   & 1.00   & 9.7   & 0.71 & 14.2 \\
& SAC           & -0.18  & 973.8  & 1.00   & 10.6  & 0.60 & 24.5 \\
& TD3           & -0.12  & 712.8  & 1.00   & 9.0   & 1.00 & 12.0 \\
& TRPO          & -0.12  & 267.3  & 1.00   & 9.0   & 1.00 & 12.0 \\
& SAC + HER     & -0.15  & 1097.4 & 1.00   & 9.2   & 0.80 & 12.1 \\
& TD3 + HER     & -0.20  & 984.1  & 0.80   & 9.0   & 0.70 & 12.3 \\
& Random policy & -2.98  & 134.1  & 0.00   & N/A   & 0.00 & N/A \\
\midrule
\multirow{10}{*}{\shortstack[l]{Env2  \\ (simulation \\ + random goal)}}      
& A2C           & -1.49  & 72.3   & 0.03   & 30.8   & 0.00 & N/A  \\
& ACKTR         & -1.64  & 74.2   & 0.04   & 48.1   & 0.00 & N/A  \\
& DDPG          & -1.94  & 523.2  & 0.02   & 39.8   & 0.00 & N/A  \\
& PPO           & -2.54  & 68.1   & 0.04   & 31.7   & 0.00 & N/A  \\
& SAC           & -2.59  & 1063.9 & 0.02   & 49.7   & 0.00 & N/A  \\
& TD3           & -2.19  & 783.7  & 0.03   & 34.9   & 0.00 & N/A  \\
& TRPO          & -1.43  & 191.8  & 0.05   & 25.7   & 0.00 & N/A  \\
& SAC + HER     & -0.27  & 1285.7 & 0.67   & 11.4   & 0.10 & 27.2 \\
& TD3 + HER     & -0.32  & 980.7  & 0.64   & 11.1   & 0.17 & 19.3 \\
& Random policy & -1.68  & 132.6  & 0.05   & 22.0   & 0.00 & N/A  \\
\midrule
\multirow{10}{*}{\shortstack[l]{Env3  \\ (physical \\ robot \\ + fixed goal)}}   
& A2C           & -1.08  & N/A    & 0.00   & N/A    & 0.00 & N/A \\
& ACKTR         & -0.28  & N/A    & 0.95   & 13.0   & 0.00 & N/A \\
& DDPG          & -0.71  & N/A    & 0.00   & N/A    & 0.00 & N/A \\
& PPO           & -0.25  & N/A    & 1.00   & 16.9   & 0.00 & N/A \\
& SAC           & -0.33  & N/A    & 0.35   & 10.0   & 0.00 & N/A \\
& TD3           & -0.19  & N/A    & 1.00   & 10.0   & 0.00 & N/A \\
& TRPO          & -0.53  & N/A    & 0.00   & N/A    & 0.00 & N/A \\
& SAC + HER     & -0.68  & N/A    & 0.00   & N/A    & 0.00 & N/A \\
& TD3 + HER     & -0.28  & N/A    & 1.00   & 10.0   & 0.00 & N/A \\
& Random policy & -2.19  & N/A    & 0.00   & N/A    & 0.00 & N/A \\
\midrule
\multirow{10}{*}{\shortstack[l]{Env4  \\ (physical \\ robot \\ + random goal)}} 
& A2C           & -1.29  &  N/A   & 0.10   & 13.0   & 0.00 & N/A  \\
& ACKTR         & -1.82  &  N/A   & 0.05   & 35.0   & 0.00 & N/A  \\
& DDPG          & -1.55  &  N/A   & 0.05   & 27.0   & 0.00 & N/A  \\
& PPO           & -1.31  &  N/A   & 0.00   & N/A    & 0.00 & N/A  \\
& SAC           & -1.43  &  N/A   & 0.00   & N/A    & 0.00 & N/A  \\
& TD3           & -1.63  &  N/A   & 0.00   & N/A    & 0.00 & N/A  \\
& TRPO          & -1.76  &  N/A   & 0.00   & N/A    & 0.00 & N/A  \\
& SAC + HER     & -0.39  &  N/A   & 0.50   & 17.9   & 0.05 & 9.0  \\
& TD3 + HER     & -0.28  &  N/A   & 0.75   & 18.5   & 0.20 & 17.5 \\
& Random policy & -1.57  &  N/A   & 0.00   & N/A    & 0.00 & N/A  \\
\bottomrule
\end{tabular}
\label{tab:raw_results}
\end{table*}

\begin{table*}[!htbp]
\centering
\caption{Evaluation metrics of all trained agents for the four environments (continued)}
\begin{tabular}{p{24mm}|p{24mm}|p{20mm}p{20mm}p{20mm}p{20mm}}
\toprule
Environment & Algorithm & Success ratio @10 mm  & Reach time @10 mm & Success ratio @5 mm & Reach time @5 mm \\
\midrule
\multirow{10}{*}{\shortstack[l]{Env1  \\ (simulation \\ + fixed goal)}}      
& A2C           & 0.04  & 63.8 & 0.01  & 91.8   \\
& ACKTR         & 0.73  & 22.1 & 0.20  & 43.3   \\
& DDPG          & 0.20  & 12.5 & 0.00  & N/A    \\
& PPO           & 0.41  & 17.1 & 0.08  & 40.3   \\
& SAC           & 0.10  & 14.0 & 0.10  & 15.0   \\
& TD3           & 1.00  & 12.5 & 0.50  & 18.2   \\
& TRPO          & 0.81  & 12.4 & 0.22  & 38.5   \\
& SAC + HER     & 0.50  & 12.6 & 0.00  & N/A    \\
& TD3 + HER     & 0.10  & 13.0 & 0.00  & N/A    \\
& Random policy & 0.00  & N/A  & 0.00  & N/A    \\
\midrule
\multirow{10}{*}{\shortstack[l]{Env2  \\ (simulation \\ + random goal)}}       
& A2C           & 0.00  & N/A  & 0.00  & N/A     \\
& ACKTR         & 0.00  & N/A  & 0.00  & N/A     \\
& DDPG          & 0.00  & N/A  & 0.00  & N/A     \\
& PPO           & 0.00  & N/A  & 0.00  & N/A     \\
& SAC           & 0.00  & N/A  & 0.00  & N/A     \\
& TD3           & 0.00  & N/A  & 0.00  & N/A     \\
& TRPO          & 0.00  & N/A  & 0.00  & N/A     \\
& SAC + HER     & 0.01  & 37.1 & 0.00  & N/A     \\
& TD3 + HER     & 0.03  & 25.9 & 0.01  & 28.6    \\
& Random policy & 0.00  & N/A  & 0.00  & N/A     \\
\midrule
\multirow{10}{*}{\shortstack[l]{Env3  \\ (physical \\ robot \\ + fixed goal)}}
& A2C           & 0.00 & N/A & 0.00  & N/A  \\
& ACKTR         & 0.00 & N/A & 0.00  & N/A  \\
& DDPG          & 0.00 & N/A & 0.00  & N/A  \\
& PPO           & 0.00 & N/A & 0.00  & N/A  \\
& SAC           & 0.00 & N/A & 0.00  & N/A  \\
& TD3           & 0.00 & N/A & 0.00  & N/A  \\
& TRPO          & 0.00 & N/A & 0.00  & N/A  \\
& SAC + HER     & 0.00 & N/A & 0.00  & N/A  \\
& TD3 + HER     & 0.00 & N/A & 0.00  & N/A  \\
& Random policy & 0.00 & N/A & 0.00  & N/A  \\
\midrule
\multirow{10}{*}{\shortstack[l]{Env4  \\ (physical \\ robot \\ + random goal)}}  
& A2C           & 0.00  & N/A  & 0.00  & N/A  \\
& ACKTR         & 0.00  & N/A  & 0.00  & N/A  \\
& DDPG          & 0.00  & N/A  & 0.00  & N/A  \\
& PPO           & 0.00  & N/A  & 0.00  & N/A  \\
& SAC           & 0.00  & N/A  & 0.00  & N/A  \\
& TD3           & 0.00  & N/A  & 0.00  & N/A  \\
& TRPO          & 0.00  & N/A  & 0.00  & N/A  \\
& SAC + HER     & 0.00  & N/A  & 0.00  & N/A  \\
& TD3 + HER     & 0.00  & N/A  & 0.00  & N/A  \\
& Random policy & 0.00  & N/A  & 0.00  & N/A  \\
\bottomrule
\end{tabular}
\label{tab:raw_results2}
\end{table*}

\section{Conclusion and Future Work}  \label{conclusion}

A benchmark procedure is described in the present paper to compare the performance of various model-free RL algorithms at solving the reaching task with a robot manipulator. It is found that it is generally more challenging to solve a reaching task where the target coordinates are initialised randomly at the beginning of each episode. In a fixed goal setting, the highest performance in terms of average return, success ratio and reach time is achieved by TD3 and TRPO. The highest sample efficiency is achieved by SAC, TRPO and DDPG. Augmenting the intrinsic reward signal of off-policy algorithms such as SAC and TD3 using the HER exploration strategy tends to destabilise slightly the training -- both in terms of repeatability and sample efficiency; however it also allows the agent to learn an efficient policy in environments where the goal is initialised randomly. Transferring policies from virtual to physical environments proved to be challenging, however similar conclusions were drawn as to which algorithms perform best. Combining TD3 with HER proved to be the most efficient strategy in physical environments.

Although the methodology described here provides a systematic and reproducible experimental procedure, the authors acknowledge that the benchmark results could be further improved by additional work. 

Firstly, some algorithms failed to solve the reaching task completely when the goal is initialised randomly, despite the dense reward setting, see Fig.  \ref{fig:learning_curves_env1}. This is likely due to failing to identify optimal hyperparameters for the configuration in question. Selecting appropriate hyperparameters for training RL agents is notoriously difficult \cite{Haarnoja2018}. A more exhaustive hyperparameter search may improve the performance. Additionally, scaling the environment's action space to lie in the interval [-1, 1] could help with the training convergence.

Secondly, the policy transfer from the simulated to the physical environment appears to be challenging as policies often fail to generalise to the physical system (also known as sim-to-real). By nature, there are inherent discrepancies between the dynamics of a simulator and the real-world. In particular, a physical robotic arm is subject to the following noise sources and non-stationarities: a) the position and speed resolution inherent to the Dynamixel servo actuators used by the WidowX MKII; and b) overheating of the actuators and wear of the body parts. These physical imperfections tend to increase the compliance margin between the desired goal position and the actual position. Performing a rigorous calibration of the physical robot would ensure that the arm's dynamics in the simulator closely match that of the real world, thus reducing this reality gap.

Another possible solution to account for these irregularities consists in introducing noise in the simulator by altering joint friction coefficients or link masses for example, thus effectively creating a collection of imperfect training environments. The most realistic virtual environment can be identified by training RL agents on these imperfect environments, deploying the learnt policies on the physical robot and selecting the one with the smallest observed difference.

The discrepancies observed between virtual and physical environments can be further reduced by executing part of the training on the physical environment in order to adjust the policy to the real world. In addition, the robot's accuracy could be further increased by restricting the joint's amplitude motion between each timestep (i.e. reducing the range of the action space).

Lastly, the complexity of the task to solve could be further increased so that the algorithm's performance can be compared in more detail. Such increases in complexity could be brought by considering a sparse reward setting, where the reward is only incremented if the end effector is positioned close enough to the goal position. Solving such a task would greatly benefit from exploration enhancement techniques such as HER or curiosity-driven exploration \cite{Pathak2017} where the intrinsic reward signal coming from the environment is augmented. The task difficulty may also be increased by defining a more challenging reward function. For example, the orientation of the arm's end effector may be included in the reward definition. Additionally, obstacles could be included in the environment with a reward penalty applied at each collision. Finally, the environment's complexity could be increased by allowing the target to move randomly at each timestep for instance.

\section*{Acknowledgements}

This Career-FIT project has received funding from the European Union’s Horizon 2020 research and innovation programme under the Marie Skłodowska-Curie grant agreement No. 713654.

\FloatBarrier

\bibliographystyle{spmpsci_unsrt}
\bibliography{library.bib}

\end{document}